\documentclass[pdflatex,sn-mathphys-num]{sn-jnl}

\usepackage{graphicx}%
\usepackage{multirow}%
\usepackage{amsmath,amssymb,amsfonts}%
\usepackage{amsthm}%
\usepackage{mathrsfs}%
\usepackage[title]{appendix}%
\usepackage{xcolor}%
\usepackage{textcomp}%
\usepackage{manyfoot}%
\usepackage{booktabs}%
\usepackage{colortbl}
\usepackage{algorithm}%
\usepackage{algorithmicx}%
\usepackage{algpseudocode}%
\usepackage{listings}%

\theoremstyle{thmstyleone}%

\theoremstyle{thmstyletwo}%

\theoremstyle{thmstylethree}%

\begin{document}

\title[Article Title]{Layer-Guided UAV Tracking: Enhancing Efficiency and Occlusion Robustness}

\author[1]{\fnm{Yang} \sur{Zhou}}\email{242260509@st.usst.edu.cn}
\author*[1]{\fnm{Derui} \sur{Ding}}\email{deruiding2010@usst.edu.cn}
\author[1]{\fnm{Ran} \sur{Sun}}\email{sunr1990@usst.edu.cn}
\author[2]{\fnm{Ying} \sur{Sun}}\email{yingsun1991@usst.edu.cn}
\author[1]{\fnm{Haohua} \sur{Zhang}}\email{233370882@st.usst.edu.cn}

\affil[1]{\orgdiv{Department of Control Science and Engineering}, \orgname{University of Shanghai for Science and Technology}, \orgaddress{\city{Shanghai}, \postcode{200093},  \country{P.~R.~China}}}

\affil[2]{\orgdiv{Busness School}, \orgname{University of Shanghai for Science and Technology}, \orgaddress{ \city{Shanghai}, \postcode{200093},  \country{P.~R.~China}}}


\abstract{Visual object tracking (VOT) plays a pivotal role in unmanned aerial vehicle (UAV) applications. Addressing the trade-off between accuracy and efficiency, especially under challenging conditions like unpredictable occlusion, remains a significant challenge. This paper introduces LGTrack, a unified UAV tracking framework that integrates dynamic layer selection, efficient feature enhancement, and robust representation learning for occlusions. By employing a novel lightweight Global-Grouped Coordinate Attention (GGCA) module, LGTrack captures long-range dependencies and global contexts, enhancing feature discriminability with minimal computational overhead. Additionally, a lightweight Similarity-Guided Layer Adaptation (SGLA) module replaces knowledge distillation, achieving an optimal balance between tracking precision and inference efficiency. Experiments on three datasets demonstrate LGTrack's state-of-the-art real-time speed (258.7 FPS on UAVDT) while maintaining competitive tracking accuracy (82.8\% precision). The code is available at https://github.com/XiaoMoc/LGTrack}

\keywords{ Unmanned aerial vehicle tracking, Vision Transformer, Global-grouped coordinate attention, Similarity-guided layer adaptation, Occlusion.}



\maketitle

\section{Introduction}\label{sec1}

Visual Object Tracking (VOT)~\cite{Aklak_2023,Gui_2024,Zhang_2025} is a fundamental task in computer vision, especially in the field of Unmanned Aerial Vehicle (UAV) \cite{Yu_2026,Nan_2025}. In recent years, Unmanned Aerial Vehicle (UAV) tracking, as an important branch of VOT, has enabled a wide range of real-world applications such as intelligent transportation and autonomous navigation and decision-making~\cite{gharsa2025autonomous}. In the comparison with ground-level tracking, UAV tracking is extremely challenging due mainly to the aerial top-down viewpoint, which induces severe cluttered background, drastic target scale variations, and frequent target occlusions. Furthermore, UAV platforms are constrained by limited onboard computing resources and energy budgets. As such, it is necessary to achieve high tracking accuracy under strict real-time latency requirements. Therefore, an effective UAV tracker realize the balance between precision and inference speed. Benefiting from the efficiency of Discriminative Correlation Filters (DCFs), DCF-based  trackers have been widely adopted in UAV scenarios; but the reliance on hand-crafted features often limits their robustness under complex appearance changes and cluttered backgrounds \cite{danelljan2015learning,danelljan2017discriminative,henriques2014high}. Siamese Network (SN)-based trackers provide stronger discrimination than DCF-based ones by introducing deep feature learning. It should be pointed out that the adopted designs via Convolutional Neural Network (CNN) still exhibit shortcomings due to limited global context modeling, thereby leaving room for further improvement in accuracy \cite{IJNDI2025Shao,li2018high,wu2025learning}. Recently, Vision Transformers (ViTs), empowered by global self-attention, have become mainstream backbones for VOT tasks. Representative one-stream trackers, such as MixFormer \cite{cui2022mixformer} and OSTrack \cite{ye2022joint}, realize both feature extraction and template-search interaction within a single encoder. Such a strategy not only enhances the parallelism capability but also improves tracking accuracy. However, standard ViT architectures are often computationally expensive and parameter-heavy, such that the direct deployment on UAV platforms is very difficult.

To improve efficiency, a series of lightweight ViT-based trackers has been explored in recent years. For example, Aba-ViTrack \cite{li2023adaptive} introduces an adaptive termination strategy of
background-aware token to reduce redundant computation. AVTrack~\cite{wu2025learning} further proposes a dynamic layer activation rule based on input complexity to increase efficiency while avoiding unstructured sparsity. Unfortunately, the adoption of auxiliary classifiers for activation decisions typically results in temporal cost and structural redundancy. To mitigate layer redundancy, SGLATrack \cite{xue2025similarity} employs a lightweight selection module to identify and retain only the most representative single layer for feature encoding, and introduces an inter-layer similarity loss to guide stable layer selection. This line of work indicates that a large portion of ViT depth can be redundant for tracking, and that dynamic layer selection provides a promising direction for real-time UAV deployment.

\begin{figure}[t]
	\centering
	\includegraphics[width=0.8\textwidth]{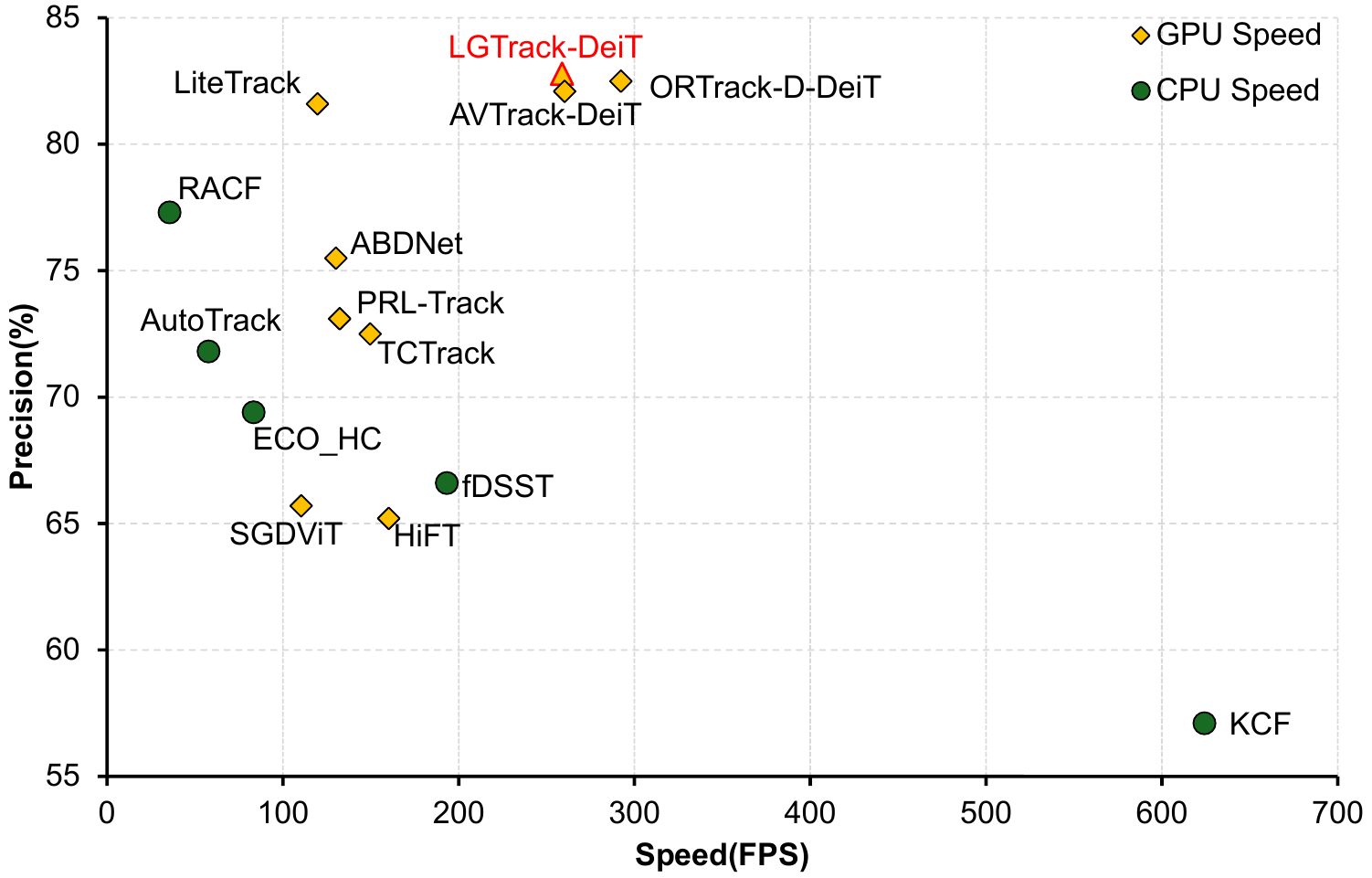}
	\caption{Compared to SOTA UAV trackers on UAVDT, our LGTrack-DeiT achieves a record with 82.8\% precision and a speed of 258.7 FPS, where ORTrack-DeiT relies on knowledge distillation.}
	\label{fig_Prec_Spd}
\end{figure}

Despite encouraging progress in lightweight ViT design, existing one-stream ViT trackers still exhibit limited robustness under frequent and complex occlusion scenarios in UAV videos, such as building occlusions and vegetation interference. When the target is partially or fully occluded, the feature correspondence between template and search regions can degrade rapidly, often leading to tracking drift or failure. For instance, ORTrack \cite{wu2025learningorr} addresses this challenge by leveraging a spatial Cox process to stochastically mask target regions during training, thereby simulating realistic occlusions, and applies feature-consistency constraints to encourage occlusion-invariant representations without introducing inference overhead. It should be pointed out that ORTrack additionally adopts knowledge distillation to balance efficiency and performance. However, distillation can introduce complexities to the training process, often extending training time, and may also impose a performance ceiling that depends on the teacher model. In contrast, our paper takes a more direct and architecture-focused approach to enhance efficiency. Specifically, by dynamically selecting a single representative layer for feature encoding, our tracker reduces computation burden without external supervision, thus maintaining a simpler training pipeline and achieving the trade-off between accuracy and inference speed. It follows from Fig.~\ref{fig_Prec_Spd} that the processing speed of our tracker is slightly lower than that of ORTrack.

According to the occlusion simulation during the training process, we believe that the enhancement of representational quality for core feature maps will be a more direct and effective approach to handle occlusion and background clutter. This perspective has been well verified by recent progress in conditional generative modeling and controllable synthesis. In these fields, the enhancement of feature controllability and the improvement of representation alignment are key factors for achieving robust performance under challenging conditions \cite{shen2024imagpose}. SCNet \cite{lin2025scnet} combines the Swin Transformer branch for global context modeling with the ConvNeXt branch for fine-grained local texture extraction, and also achieves adaptive fusion through feature fusion modules equipped with space-channel attention. Although SCNet is designed for image restoration rather than visual tracking, its successful approach to coordinating global semantics and local details offers valuable inspiration for feature enhancement in drone tracking. Motivated by these results, we propose a Global Grouped Coordinate Attention (GGCA) module, which is inserted between the ViT encoder and the prediction head. GGCA takes the ViT feature maps as inputs and applies grouped coordinate attention to capture long-range dependencies via explicit positional encoding along both height and width directions. Meanwhile, it further enhances the complementary of global contexts through dual pathways of average pooling and max pooling. The resulting attention weights recalibrate the channel and spatial responses, which helps enhance target-related cues while effectively suppressing irrelevant background interference. As a lightweight plug-and-play component, GGCA improves robustness to both occlusion and deformation without substantial inference cost.

In summary, this paper proposes \textit{a novel tracking framework}, LGTrack,  to enhance the efficiency and robustness of UAV tracking through the co-design of three aspects around ViT backbones. In the depth aspect, the Similarity Guided Layer Adaptation (SGLA) module is provided to dynamically select the optimal layer based on inter-layer feature similarity, such that LGTrack can bypass redundant layers. In the feature aspect, the proposed GGCA module enhances the final output features of ViT, strengthening the feature discrimination under occlusions and backgrounds. In the training aspect, occlusion-simulated learning is creatively adopted to encourage occlusion-robust feature representations. Extensive experiments on three popular datasets demonstrate that LGTrack achieves a superior accuracy-speed trade-off. Notably, as shown in Fig. \ref{fig_Prec_Spd}, LGTrack achieves 82.8\% precision while running at 258.7 FPS on UAVDT. Our main contributions are summarized as follows.
\begin{itemize}
	\item A novel lightweight plug-and-play module, GGCA, is constructed to capture global context and long-range dependencies, enhancing feature discriminability against clutter with minimal inference overhead.
	\item A lightweight SGLA module is employed to replace distillation in ORTrack, eliminating the performance ceiling limitations of teacher networks and the time- and computation-intensive knowledge distillation process during training, while achieving an optimal balance between tracking precision and inference efficiency. 
	\item LGTrack, a unified UAV tracking framework based on a ViT backbone, is designed in light of the above two modules. Experiments on three datasets validate that LGTrack delivers a strong accuracy-speed trade-off and achieves state-of-the-art real-time performance in UAV tracking.
\end{itemize}

\section{Related Work}\label{sec:rw}

This section briefly introduces the development of visual tracking methods, with a particular focus on ViT-based results, and then discusses some important designs in occlusion-robust feature representation and coordinate attention, thereby highlighting the motivation of our work.

\subsection{Visual Object Tracking } 

Currently, mainstream VOT methods can be broadly categorized into three classes: DCF-based methods, SN-based methods, and ViT-based methods. DCF-based trackers are widely utilized in drone tracking due to their high efficiency on CPUs. However, their handcrafted architecture results in poor feature representation capability, leading to low robustness in high-noise environments \cite{IJNDI2025Qiang,Danelljan2016ECOEC,danelljan2017discriminative,henriques2014high,li2020autotrack}. SN-based trackers achieve higher accuracy and robustness than DCF methods by incorporating the advantages of CNNs, but they generally require significant computational resources compared to DCF trackers, resulting in a lack of efficiency \cite{IJNDI2025Chen,li2018high,xu2020siamfc++}. Recently, increasing attention has focused on developing ViT-based trackers for generic visual tracking \cite{hu2023transformer, shi2024evptrack, xie2024autoregressive}. For example, Mixformer \cite{cui2022mixformer} achieves deep fusion of template and search features via iterative mixed attention and a single-stream architecture, while OSTrack \cite{ye2022joint} enhances the ability to distinguish similar targets through joint feature learning and relationship modeling. Nevertheless, although ViT-based trackers have advanced beyond earlier CNN-based methods in both exploration depth and accuracy, they remain computationally intensive, heavily dependent on high-performance hardware, and inherently cumbersome, thereby struggling to meet real-time tracking requirements. 

A promising strategy of efficiency improvement is to optimize ViTs through conditional computation. This strategy dynamically allocates computation resources based on the complexity of input samples. The related results can be divided into two categories: adaptive token pruning \cite{yin2022vit} and adaptive layer activation \cite{bakhtiarnia2021multi,wu2025learning}. In the first category, AViT \cite{yin2022vit} introduces an adaptive token refinement rule, in which redundant background tokens are discarded according to a dynamic stop probability. Similarly, Aba-ViTrack \cite{li2023adaptive} reduces redundant computation through adaptive token calculation. This approach enhances the ability to filter out background interference while maintaining tracking accuracy. However, these trackers could perform memory access operations that are unoptimized and unstructured because of dynamic changes in the number of tokens. As a result, there is room for improvement in the trade-off between accuracy and efficiency. The idea of the second category is to adaptively activate ViT layers depending on input complexity. For example, ViT-EE \cite{bakhtiarnia2021multi} deploys internal classifiers after each layer to decide whether to terminate model inference early by using a dynamic logic based on layer-by-layer confidence evaluation. AVTrack \cite{wu2025learning} optimizes computational efficiency by dynamically activating Transformer blocks via a dedicated module and hence avoids the redundant calculations of full-block activation. It assesses input complexity and then skips ViT layers based on the output probability of an internal module. SGLATrack \cite{xue2025similarity} proposes an SGLA method to accelerate the ViT structure via observations of layer redundancy. While SGLATrack focuses mainly on structural acceleration, our approach integrates layer selection and occlusion-robust learning into a unified framework for joint optimization. Specifically, we introduce an SGLA module to dynamically deactivate layers with similar representations and keep only a single optimal layer, and therefore achieve a better trade-off between accuracy and inference speed.

\subsection{Occlusion-Robust Feature Representation}

Occlusion-robust feature representation plays an important role in computer vision and image processing. It enables models to reliably recognize and process targets even under partial occlusion \cite{park2022handoccnet}. Early research primarily employed handcrafted features and improved the model performance in certain scenarios. However, the robustness is still insufficient when dealing with the complex and variable challenges of real-world environments. The advance of deep learning has popularized the use of CNNs and related architectures for extracting occlusion-robust representations \cite{jiang2024occlusion,park2022handoccnet}. Deep networks are capable of modeling intricate patterns and adapting to occlusion scenarios, thereby yielding more robust features. Such methods demonstrate considerable advantages in various tasks such as object detection\cite{chi2020pedhunter}, action recognition \cite{das2024occlusion}, and visual tracking \cite{askar2020occlusion}. However, there are still significant shortcomings in the field of drone tracking systems. Very recently, a random masking strategy based on spatial Cox process modeling has been provided in ORTrack \cite{wu2025learningorr} to enhance the invariance of target feature representations. Yet, its dependence on knowledge distillation for efficiency improvement introduces an additional training phase, and hence, the computational cost has increased in comparison with direct student model training. In this work, we propose a stochastic masking method to learn occlusion-robust feature representations via a spatial Cox process. Furthermore, to improve inference efficiency, we also integrate a selection module that dynamically disables redundant layers. Unlike ORTrack \cite{wu2025learningorr}, our approach removes the process of teacher-student distillation and adopts a single-network framework to realize the balance between occlusion robustness and inference efficiency in drone tracking.

\subsection{Coordinate Attention}

The core idea of attention mechanisms is to simulate the human visual system such that neural networks can adaptively focus on key information of input features while minimizing distractions. This approach has become a cornerstone for enhancing the representational capability of deep learning models. This mechanism is dedicated to modeling the dependency across channel and spatial feature information more efficiently and accurately. The early channel attention module, SENet \cite{hu2018squeeze}, captures global context via global average pooling and generates channel weights by using fully connected layers. However, it overlooks the crucial spatial location information. To overcome this shortcoming, subsequent achievements such as CBAM \cite{woo2018cbam} obtain the relationship between channel and spatial feature information by introducing hybrid attention mechanisms. However, these methods often rely on standard convolutional kernels to aggregate spatial features, which results in a limited capture capability of long-range pixel relationships and increased computational load. By resorting to one-dimensional encoding along height and width directions, Coordinate Attention (CA) \cite{hou2021coordinate} offers an elegant solution to capture long-range and cross-channel dependencies while embedding precise position information into the attention maps. It should be pointed out that intermediate layers require a significant amount of computational effort for features with high channel dimensions, and the uniform treatment mode of all channels may limit the diversity of learned attention patterns. To overcome these challenges, we introduce the GGCA module. Its key innovation lies in a grouped structure, which divides input channels into multiple subgroups and performs CA on each subgroup independently. This design enables different groups to learn diverse attention features, thereby significantly enhancing the expression ability of the model with very little speed cost. Additionally, we aggregate contexts by adopting a dual-path pooling strategy, which combines global average pooling and global max pooling. It enables a more comprehensive capture of the statistical representations from feature maps and enhances the robustness to complex visual patterns. In occlusion scenarios, the target's visible parts often provide crucial information. Global maximum pooling enhances the features of these unobstructed significant areas, while global average pooling provides background context information to helpfully infer the target's occlusion area from a global perspective. These complementary characteristics enable the developed model to maintain robust recognition and localization ability even if the target information is incomplete. Benefiting from the location-aware advantages of CA as well as the grouped and dual-path design, GGCA achieves feature enhancement while guaranteeing a better balance between effectiveness and computational efficiency.

\begin{figure*}[t]
	\centering	\includegraphics[width=1.0\textwidth]{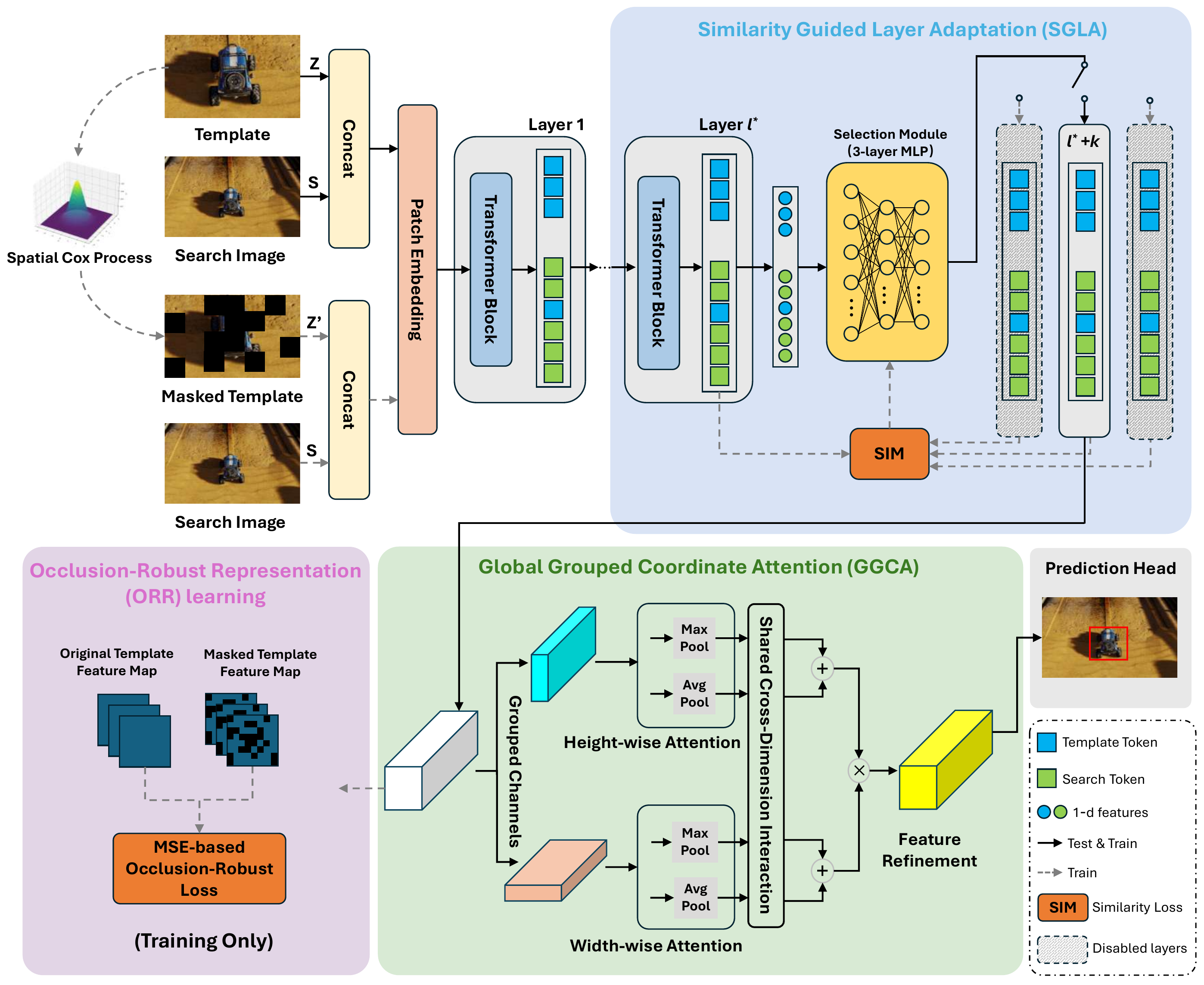}
	\caption{Overview of the proposed LGTrack framework based on ViT backbones, including $L$ Transformer blocks. It involves two core co-designed modules (i.e. SGLA and GGCA ) for feature refinement, and a robust learning (i.e. ORR learning) strategy for training. During training, the selection module (i.e. the 3-layer MLP) is determined by using the proposed layer-wise similarity loss function. During inference, GGCA captures long-range dependencies and global contexts, and the 3-layer MLP module in SGLA activates the optimal layer in the last $L-l^\ast$ Transformer blocks and disables redundant layers to avoid waste of computing resources.} \label{fig:overview}
\end{figure*}

\section{Proposed Method}\label{sec:method} 

In this section, we first present an overview of our proposed real-time UAV tracking framework, LGTrack, as illustrated in Fig.~\ref{fig:overview}. Then, we detail two core co-designed modules, that is, SGLA for efficient inference and GGCA for enhanced feature representation. Furthermore, we describe the robust learning strategy via occlusion-robust representations based on spatial Cox processes employed in ORTrack.

\subsection{Overview}
This section discusses the main structure of our ViT tracking framework. As shown in Fig.~\ref{fig:overview}, the proposed LGTrack framework based on a ViT backbone includes two modules, SGLA for efficient inference and GGCA for enhanced feature representation, as well as a robust learning strategy based on ORR and layer-wise similarity loss. In what follows, let us concretize the initial input of LGTrack and the outputs of the corresponding modules.

The inputs of our LGTrack includes a template image ${Z} \in \mathbb{R}^{3\times H_z \times W_z}$ and a search image ${S} \in \mathbb{R}^{3\times H_s \times W_s}$. They are transformed into a template token sequence ${X}_z \in \mathbb{R}^{N_z \times D} $ and a search token sequence ${X}_s \in \mathbb{R}^{N_s \times D} $ via a patch embedding layer. The joint token sequence can be represented as
\begin{equation}
	{X}=[{X}_z,{X}_s] = \mathcal{E}({Z}, {S}) \in \mathbb{R}^{N \times D},
	\label{eq:PElayer}
\end{equation}
where $D$ is the embedding dimension, $[\cdot]$ denotes the concatenation operation, and $N=N_z+N_s$ is the total number of tokens. 

Different from ORTrack, our LGTrack framework includes $L$ Transformer blocks, where the first $l^\ast$ Transformer blocks are fixed and only one block in the last $L-l^\ast$ ones is activated via the 3-layer MLP module in SGLA during inference. Let $\mathcal{T}^i$ represent the $i$-th Transformer layer and its feature update can be described by ${X}^i=\mathcal{T}^{i}({X}^{i-1})$. Furthermore, the forward computation process of ViT can be formulated as
\begin{equation}
	{X}^{l^\ast+k} =[{X}^{l^\ast+k}_z,{X}^{l^\ast+k}_s]= \mathcal{T}^{l^\ast+k} \circ \mathcal{T}^{l^\ast} \dots \circ \mathcal{T}^1 \circ \mathcal{E}({Z}, {S}),
	\label{eq:ViT}
\end{equation}
where ${X}^{l^\ast+k}$ is the final output feature, and $\circ$ denotes composition operation. Furthermore, ${X}^{l^\ast+k}_z$ means the template feature and ${X}^{l^\ast+k}_s$ is the search region feature.

In what follows, let us further discuss the activation via the 3-layer MLP module in SGLA. 
To optimize the computational efficiency of the ViT encoder, empirical evidence suggests that some deep layers are redundant and can be disabled without significant accuracy degradation. In object-tracking tasks, more discriminative information often comes from shallow layers that capture detailed features, rather than from deep layers that encode high-level semantics. In the comparison with ORTrack, we employ a dynamic layer selection mechanism (i.e. the 3-layer MLP module) in the lightweight ViT architecture. This mechanism aims to select the most representative single-layer output from $L-l^\ast$ Transformer blocks as the final feature.

After obtaining the output of the ViT encoder, we introduce a novel GGCA module to enhance the search region features. It should be pointed out that ViT is essentially based on global semantic dependencies and hence lacks explicit spatial localization capabilities. In light of the advantages of convolution, the generated attention weights via GGCA can accurately focus on target regions and suppress background noise, facilitating the feature map's ability to possess both global semantic correlations and identify the target area and its surrounding local spatial range. Specifically, the search region tokens $X_s^{l^\ast+k}\in\mathbb{R}^{N_s\times D}$ are reshaped into a 2D spatial feature map $F_g\in\mathbb{R}^{C\times H\times W}$ , which is then refined by the GGCA module
\begin{equation}
	F_g'=\mathrm{GGCA}(F_g),
\end{equation}
where $F_g'\in\mathbb{R}^{C\times H\times W}$ is the enhanced feature map. Finally, the target bounding box is computed by a prediction head based on the enhanced features ${B}=\mathcal{H}({F_g'})$.

During the training phase, inspired by ORTrack, we also incorporate the target template  $Z'=\mathfrak{m}(Z)$ after random mask, where $\mathfrak{m}(\cdot)$ denotes a structured random masking operation based on a spatial Cox process, masking the non-overlapping image blocks of a size of $b\times b$ with a specific mask ratio. This process generates masking patterns that are more likely to concentrate near the target center, thereby simulating a wider range of realistic occlusion scenarios. To obtain occlusion-robust representations, we minimize the Mean Squared Errors (MSEs) between the representations of this randomly masked template and the above unmasked one.

The above presentation roughly demonstrates the process of inference and training of the proposed LGTrack framework. Below, we will detail SGLA and GGCA, as well as the robust learning approach based on ORR.

\subsection{Global Grouped Coordinate Attention }  
\begin{figure*}[t]
	\centering	\includegraphics[width=1.0\textwidth]{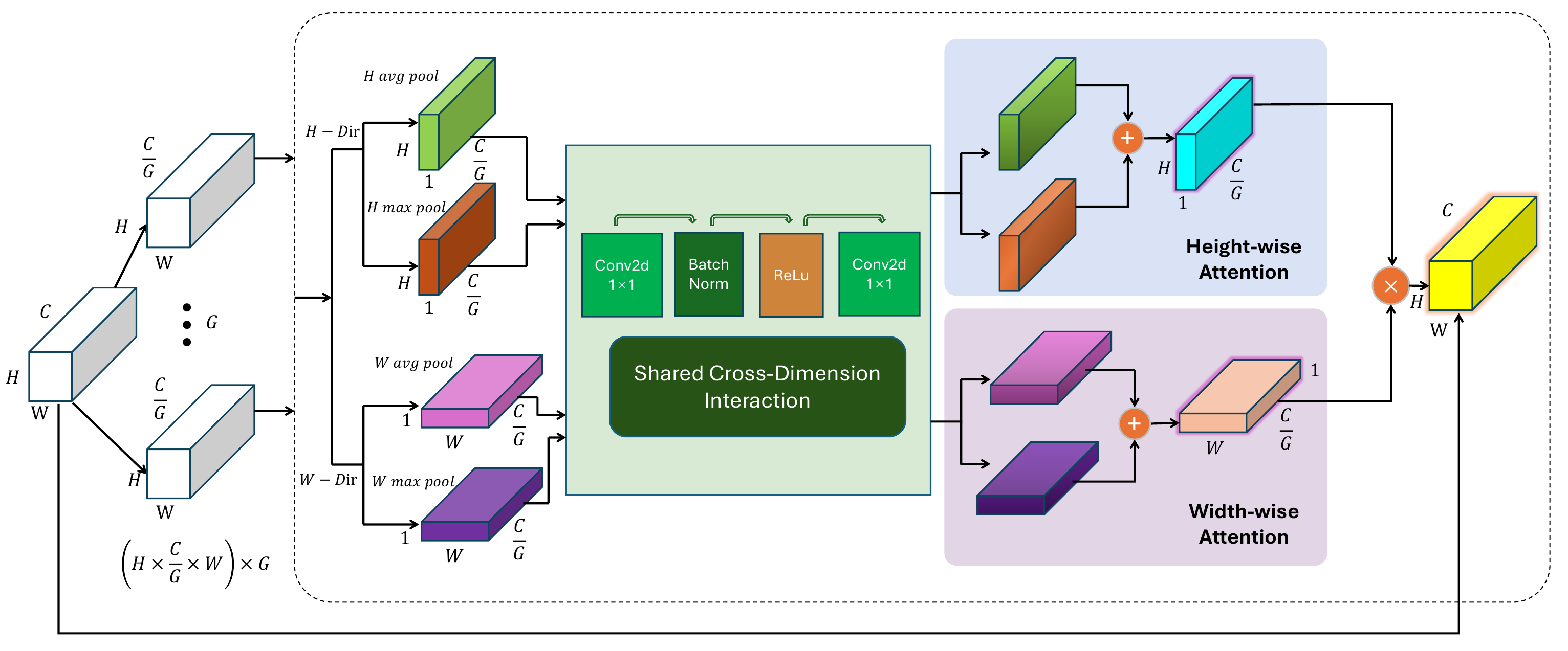}
	\caption{Overall architecture of GGCA, including the detailed internal structure and feature-processing pipeline, where G is the number of groups, $\oplus$ is the element-wise addition, and $\otimes$ is the dot product.} \label{fig:GGCA}
\end{figure*}

It should be pointed out that typical SENet \cite{hu2018squeeze}, CBAM \cite{woo2018cbam}, and CA \cite{hou2021coordinate} suffer from some general drawbacks, failing to strike an optimal balance among efficient global spatial-channel dependency modeling, lightweight computational overhead, and diversified attention representation learning. To overcome these shortages, a novel GGCA module is proposed in this paper, see Fig.~ \ref{fig:GGCA} for its overall architecture. GGCA leverages global information from the feature map across spatial dimensions (height and width) to generate attention maps. Furthermore, these attention maps are utilized to weight the input feature map, thereby enhancing feature representation.

For the input feature map $F_g\in\mathbb{R}^{B\times C\times H\times W}$ from SGLA modules, we divide it into $G$ groups along the channel dimension, with each group containing $C/G$ channels. Here, $B$ is the batch size, $C$ is the number of channels, and $H$ and $W$ are the height and width of the feature map, respectively. The grouped feature map is denoted as $\mathbf{X}\in\mathbb{R}^{B\times G\times\frac{C}{G}\times H\times W}$.
Next, we perform global average pooling and global max pooling on the grouped feature map along the height and width directions, respectively. Denoting $m \in \{\text{avg}, \text{max}\}$, the pooled features are
\begin{equation}
	\begin{aligned}
		\mathbf{X}_{h, m} &= \text{Pool}_m(\mathbf{X}) \in \mathbb{R}^{B \times G \times \frac{C}{G} \times H \times 1}, \\
		\mathbf{X}_{w, m} &= \text{Pool}_m(\mathbf{X}) \in \mathbb{R}^{B \times G \times 1 \times \frac{C}{G} \times W}.
	\end{aligned}
\end{equation}

In what follows, let us realize the shared cross-dimension iteration. The convolutional parameters are utilized by all groups and all pooling branches. In comparison with independent convolutions for every branch, the number of parameters is directly decreased to $1/G$. In addition, if two separate convolutional modules are employed in the interaction between the height and width dimensions, the parameters of both branches will become mutually independent. Each convolutional module is limited to learning features on a single dimension and cannot capture implicit associations between height and width dimensions. The adoption of a shared module allows different dimensional interactions to learn a unified transformation rule, leading to information complementarity between dimensions. To this end, we first apply a shared convolutional layer to process the features of each grouped feature map. This shared convolutional layer consists of two 1×1 convolutional layers, batch normalization, and a ReLU activation function, which are used to reduce and then restore the channel dimensions
\begin{equation}
	\begin{split}
		\mathbf{Y}_{h,m}=\mathrm{Conv}(\mathbf{X}_{h,m}),\;
		\mathbf{Y}_{w,m}=\mathrm{Conv}(\mathbf{X}_{w,m}).
	\end{split}
\end{equation}

By summing the outputs of the convolutional layers and applying the Sigmoid activation function, attention weights $\mathbf{A}$ are generated along the height and width directions
\begin{equation}
	\begin{split}
		\mathbf{A}_h &= \sigma(\mathbf{Y}_{h,\text{avg}} + \mathbf{Y}_{h,\text{max}}) \in \mathbb{R}^{B \times G \times \frac{C}{G} \times H \times 1}, \\
		\mathbf{A}_w &= \sigma(\mathbf{Y}_{w,\text{avg}} + \mathbf{Y}_{w,\text{max}}) \in \mathbb{R}^{B \times G \times \frac{C}{G} \times 1 \times W},
	\end{split}
\end{equation}
where $\sigma$ denotes the Sigmoid activation function. Finally, we apply the attention weights to the input feature map to generate the output feature map
\begin{equation}
	{F_g'} = \mathbf{X} \times \mathbf{A}_h \times \mathbf{A}_w \in \mathbb{R}^{B \times C \times H \times W}.
\end{equation}
Here, the attention weights $\mathbf{A}_h$ and $\mathbf{A}_w$ are expanded along the height and width directions, respectively, to match the dimensions of the input feature map.

\subsection{Similarity-Guided Layer Adaptation} 

It should be pointed out that ORTrack \cite{wu2025learningorr} requires iterative distillation between the teacher and student networks. Such a structure suffers from the performance ceiling limitations of teacher networks, besides substantial extra training time and excessive computational resource consumption due to repeated forward and backward propagation operations. When considering traditional pruning methods, directly removing these layers could result in significant performance degradation, even though deep-layer features exhibit high similarity. In order to overcome such a shortage, our approach employs a lightweight SGLA module to replace distillation in ORTrack. Specifically, we choose to retain the single most representative subsequent layer, whose output should maintain maximum similarity with the saturated layer features to preserve model performance. The main mechanism is that as network depth increases, the model's attention progressively focuses on the target region. If effective attention is already achieved at the saturated layer, the primary role of subsequent layers is to reinforce this attention and suppress background interference \cite{xue2025similarity}. Thus, selecting the subsequent layer whose output is most similar to the saturated layer helps maintain the attention mechanism's concentration and consistency. 

Let ${X}^{l^*}$ denote the saturated features of the $l^*$-th layer. The research goal is to determine the index $k$ so that the corresponding layer feature ${X}^{l^*+k}$ satisfies
\begin{equation}
	\mathop{\arg\max}\limits_{k} \; \text{Cos}({X}^{l^*},{X}^{l^*+k}),
	\label{eq:objective}
\end{equation}
where $k \leq L-l^*$ is the layer index after the saturated layer (that is, the $l^\ast$ layer). In the actual inference, the task is completed by a designed dynamic selection module $\mathcal{M}$, which adopts a multi-layer perceptron (MLP) structure. The input of the module is $\textbf{z}:= \textbf{e}^T_1 {X}^{l^*}  \in \mathbb{R}^{N}$, where $\textbf{e}^T_1 =[1,0,\ldots,0] \in \mathbb{R}^{N}$. Obviously, the selection module $\mathcal{M}$ takes the first element of the projected token as a compact global semantic agent. Benefiting from the modeling ability of inherent global contexts in ViTs, such a design can achieve accurate decision-making of layer selection without complex feature interaction. Its low structural overhead increases the speed of adaptive computation while maintaining tracking precision. The MLP then processes the input $\textbf{z}$ and outputs the selection probability distribution for each subsequent layer
\begin{equation}
	\hat{{y}}=[\hat{y}^1,\hat{y}^2,\ldots,\hat{y}^{K}]=\sigma(\mathcal{M}(\textbf{z})) \in \mathbb{R}^K,
	\label{eq:module}
\end{equation}
where, $\sigma(\cdot)$ is the sigmoid activation function. The model only retains the network layer with the largest probability value, and skips other layers. To achieve the optimization of \eqref{eq:objective}, we design an inter-layer similarity loss function $\mathcal{L}_{sim}$ to guide the training of the selection module $\mathcal{M}$
\begin{equation}
	\mathcal{L}_{sim} = \frac{1}{L-l^\ast}\sum_{k=1}^{L-l^\ast}|\hat{y}^k-y^k|.
	\label{eq:loss}
\end{equation}
The expected probability $y^k$ is defined as
\begin{equation}
	y^k = 
	\begin{cases}
		1, & \text{if } \text{Cos}({X}^{l^*},{X}^{l^*+k}) \text{ is the maximum,} \\
		0, & \text{otherwise.}
	\end{cases}
	\label{eq:loss_condition}
\end{equation}

Obviously, the selection module can adaptively configure the network structure according to the characteristics of the input samples. It significantly reduces the computation burden and improves inference speed while achieving competitive tracking accuracy. Following SGLATrack \cite{xue2025similarity}, we manually tune the hyperparameter $l^*$ to achieve the trade-off between accuracy and speed. The specific criteria for determining the saturated state will be discussed in detail in the ablation study section.

\subsection{Robust Learning via Occlusion-Robust Representations} 

Uniform random masking enhances the learn ability to robust feature representations, so that the trained model can resist interference of noises or information missing. However, the effect of this strategy is limited because it masks all spatial positions with equal probability. In real-world scenarios, the target template often contains background areas, which limit the authenticity of occlusion simulation. To address this issue, we increase the probability of masking the target's spatial positions under the fixed mask ratio, so as to improve the effectiveness of occlusion simulation. By doing so, we adopted the robust learning strategy via occlusion-robust representations based on spatial Cox processes \cite{wu2025learningorr}. 

Uniform random masking in MAE \cite{He2021MaskedAA} adopts two related random matrices $\mathbf{m}=(m_{i,j})$ and $\mathbf{b}=(b_{i,j})$, where $1\leqslant i\leqslant H_z/b$ and $1\leqslant j\leqslant W_z/b$. Here, each entry $m_{i,j}$ is independently generated via a uniform distribution on the interval $[0,1]$, and the binary variable $b_{i,j}\in \{0,1\}$ equals $1$ if and only if $m_{i,j}\in \textup{TopK}(\mathbf{m},K)$, otherwise it is 0. Here, $\textup{TopK}(\mathbf{m},K)$ is the $\textup{TopK}$ operator, which returns the set of the $K$ largest elements in $\textbf{m}$, with $K=\lfloor (1-\sigma)H_z W_z \rceil$ where $\lfloor x \rceil$ denotes the nearest integer to $x$. Mathematically, the transformation is expressed as $\mathfrak{m}_{\textup{U}}(Z)=Z\odot (\mathbf{b}\otimes \textbf{1})$, in which $\odot$ represents the Hadamard product, $\otimes$ is the tensor product, and $\textbf{1}$ is a full-ones matrix of size $b\times b$. 

The Cox process can be viewed as a two-stage stochastic mechanism and hence can be referred to as a doubly stochastic Poisson process. Specifically, the first stage is to generate the intensity function $\lambda(x,y)$. The second stage utilizes the generated $\lambda(x,y)$ to simulate a non-homogeneous Poisson point process \cite{Mattfeldt1996StochasticGA}. This paper employs the thinning algorithm \cite{Chen2016ThinningAF} to simulate the non-homogeneous Poisson point process. This algorithm first simulates a homogeneous Poisson point whose rate is higher than the maximum possible rate of the non-homogeneous process, and then thins it by removing some generated points to conform to the target intensity function. Benefiting from such an algorithm, the generated  $\lambda(x,y)$ is a bell-shaped function, thus assigning higher intensity values to the central part of the bounded region $\mathcal{B}$, which is a rectangular area with a size of $H_z/b\times W_z/b$, corresponding to the template region. If the Cox process is simulated in the reguion $\mathcal{B}$, and the resulting point pattern is denoted as $\Xi$, then a matrix $\mathbf{b}=(b'_{i,j})_{1\leqslant i\leqslant H_z/b,1\leqslant i\leqslant W_z/b}$ can be obtained, satisfying $b'_{i,j}$ equal to 1 if $(i,j)\in \Xi$, otherwise it is 0. With the help of such a matrix, the expected dropping ratio in our method is consistent with that of uniform random masking. Consequently, the adopted method is capable of simulating a wide range of occlusion patterns due to the uniform intensity distribution and stochasticity of the dropping ratio. 

Assuming that the total number of tokens is $\mathcal{K}=\mathcal{K}_{Z}+\mathcal{K}_S$, and the embedding dimension of each token is $d$. For the inputs $Z$ and $S$ through the $L$-th layer of the backbone network, all output tokens are denoted as $\mathbf{t}_{1:\mathcal{K}}^{L}(Z,S)=[\mathbf{t}_{\mathcal{K}_{Z}}^{L},\;\mathbf{t}_{\mathcal{K}_S}^{L}]\in \mathbb{R}^{\mathcal{K}\times d}$, where $\mathbf{t}_{\mathcal{K}_{Z}}^{L}$ and $\mathbf{t}_{\mathcal{K}_S}^{L}$ correspond to the tokens of the template and search images. Similarly, for inputs $Z'$ and $S$, the output tokens are $\mathbf{t}_{1:\mathcal{K}}^{L}(Z',S)$. By locating the corresponding indices in their respective token sequences, the representations of feature of $Z$ and $Z'$ can be obtained. The occlusion-aware representations can be learned by minimizing the mean squared error between the feature representations of $Z$ and $Z'$
\begin{equation}\label{Eq_MI_loss}
	\mathcal{L}_{orr}=||t_{\mathcal{K}_{Z}}^L(Z,X;\mathcal{B}_T)-t_{1:\mathcal{K}_{Z'}}^L(Z',X;\mathcal{B}_T)||^2.
\end{equation}

In the model inference phase, only $[Z,S]$ is required to be input, without performing the random template dropping operation. Therefore, this method does not introduce any additional computational overhead during inference. It is important to emphasize that this method does not depend on a specific type of ViTs, that is, any efficient ViT architecture can be applied within this framework.

\section{Experiments}\label{sec:exp} 

Following traditional protocols, our proposed LGTrack is trained on four train datasets, including LaSOT \cite{Fan2018LaSOTAH}, COCO \cite{2014Microsoft}, TrackingNet \cite{2018TrackingNet}, and GOT-10k \cite{2021GOT}. Furthermore, to validate the superiority, lots of experiments are performed in comparison with multiple state-of-the-art UAV tracking approaches on three large-scale test datasets, namely, DTB70 \cite{Li2017VisualOT}, UAVDT \cite{du2018the}, and UAV123 \cite{Mueller2016ABA}. Now, let us disclose these three datasets briefly.

\textbf{\emph{DTB70}} \cite{Li2017VisualOT} consists of 70 UAV sequences, which include various cluttered scenes and objects of different sizes, and primarily addresses the problem of severe UAV motion.
\textbf{\emph{UAVDT}} \cite{du2018the} is mainly used for vehicle tracking with various weather conditions, flying altitudes and camera views.
\textbf{\emph{UAV123}} \cite{Mueller2016ABA}, a large-scale aerial tracking dataset, involves 123 challenging sequences with more than 112K frames.

\subsection{Implementation Details}

Our tracking framework, LGTrack, is implemented in Python using PyTorch 1.12.0, with CUDA version 10.2. All experiments are conducted on the same computer with a TitanX GPU.

\emph{(1) Model}. We employ DeiT-tiny ViT as a backbone to build the tracker for extensive evaluation, i.e., LGTrack-DeiT. The number of Transformer blocks is $12$ and $l^*$ is set to $8$. The module $\mathcal{M}$ in SGLA is a 3-layer MLP, whose hidden dimension is 160. In terms of input data, we take both a template image with $128\times 128$ pixels and a search image with $256\times 256$ pixels as the input of the tracker. 

\emph{(2) Training strategy}. Following traditional protocols, we train our models on four train datasets, that is, LaSOT, COCO, TrackingNet, and GOT-10k \cite{2021GOT}. Similar to OSTrack \cite{ye2022joint}, we use typical data augmentation methods, such as horizontal flipping and brightness jittering. Furthermore, we train all trackers uniformly with 120 epochs and 60k matching pairs per epoch. With the AdamW optimizer, the learning rates are set to $4\times 10^{-4}$ for the prediction head, and $4\times 10^{-5}$ for the backbone and the selection module. The learning rates drop by a factor of $10$ after $80$ epochs. 

\emph{(3) Inference}. We follow the common practice \cite{Chen2021TransformerT,ye2022joint} and employ the Hamming window to incorporate positional priors. Specifically, a penalty is applied to the classification map by using the Hanning window. After that, the box corresponding to the highest score is selected as the prediction bounding box.

\subsection{Comparison with State-of-the-art Methods}

This study systematically evaluates the proposed LGTrack tracker and compares it with several state-of-the-art trackers, including methods based on Discriminative Correlation Filters (DCF), Convolutional Neural Networks (CNN), and Vision Transformers (ViT), see Table \ref{tab:comparision_with_light_trackers} for the experiment result. It can be observed from this table that LGTrack demonstrates outstanding overall performance, achieving competitive levels in average precision (Prec.), average success rate (Succ.), and processing speed (FPS). Compared to DCF- and CNN-based trackers, LGTrack demonstrates a significant lead in both precision and success rate while achieving real-time processing efficiency, highlighting its effectiveness and state-of-the-art capabilities for UAV tracking.

\begin{table*}[!thb]
	\scriptsize
	\centering
	\setlength\tabcolsep{5.95pt} 
	\renewcommand{\arraystretch}{1.2}
	\caption{Precision (Prec.), success rate (Succ.), and speed (FPS) of our LGTrack on three UAV tracking test datasets, i.e., DTB70 \cite{Li2017VisualOT}, UAVDT \cite{du2018the}, and UAV123 \cite{Mueller2016ABA}. {\color[HTML]{FE0000}Red}, {\color[HTML]{3531FF}blue} and {\color[HTML]{009901}green} represent the first, second and third place, respectively. Note: All Prec. and Succ. values are omitted from the percent sign (\%).}
	\label{tab:comparision_with_light_trackers}
	\begin{tabular}{ccccccccc}
		\toprule[1pt] 
		\multirow{2}{*}{} & \multirow{2}{*}{Methods} &  \multirow{2}{*}{Source} & \multicolumn{2}{c}{DTB70} & \multicolumn{2}{c}{UAVDT}  & \multicolumn{2}{c}{UAV123} \\	  
		\cmidrule(lr){4-9}	
		& & & Prec. & Succ. & Prec. & Succ.  & Prec. & Succ. \\
		\hline
		& KCF \cite{henriques2014high} & TAPMI 15 & 46.8 & 28.0 & 57.1 & 29.0 & 52.3 & 33.1 \\
		& fDSST \cite{danelljan2017discriminative} & TPAMI 17 & 53.4 & 35.7 & 66.6 & 38.3 & 58.3 & 40.5 \\
		& ECO\_HC \cite{Danelljan2016ECOEC} & CVPR 17 & 63.5 & 44.8 & 69.4 & 41.6 & 71.0 & 49.6 \\
		& AutoTrack \cite{li2020autotrack} & CVPR 20 & 71.6 & 47.8 & 71.8 & 45.0 & 68.9 & 47.2 \\
		\multirow{-5}{*}{\rotatebox{90}{DCF-based}} & RACF \cite{li2021learning} & PR 22 & 72.6 & 50.5 & 77.3 & 49.4 & 70.2 & 47.7\\ 
		\hline
		& HiFT \cite{cao2021hift} & ICCV 21 & 80.2 & 59.4 & 65.2 & 47.5 & 78.7 & 59.0 \\
		& TCTrack \cite{Cao2022TCTrackTC} & CVPR 22 & 81.2 & 62.2 & 72.5 & 53.0 & 80.0 & 60.5 \\
		& ABDNet \cite{zuo2023adversarial} & RAL 23 & 76.8 & 59.6 & 75.5 & 55.3 & 79.3 & 60.7 \\ 
		& SGDViT \cite{yao2023sgdvit} & ICRA 23 & 78.5 & 60.4 & 65.7 & 48.0 & 75.4 & 57.5 \\
		\multirow{-5}{*}{\rotatebox{90}{CNN-based}} & PRL-Track \cite{fu2024progressive} & IROS 24 & 79.5 & 60.6 & 73.1 & 53.5 & 79.1 & 59.3 \\ 
		\hline
		& LiteTrack \cite{wei2024litetrack} & ICRA 24 & 82.5 & 63.9 & 81.6 & {\color[HTML]{009901} \textbf{59.3}} & {\color[HTML]{009901} {\textbf{84.2}}} & 65.9 \\
		& SMAT \cite{gopal2024separable} & WACV 24 & 81.9 & 63.8 & 80.8 & 58.7 & 81.8 & 64.6 \\
		
		& AVTrack-DeiT \cite{wu2025learning} & ICML 24 & {\color[HTML]{FE0000} \textbf{84.3}} & {\color[HTML]{3531FF} \textbf{65.0}} & {\color[HTML]{009901} \textbf{82.1}} & 58.7 & {\color[HTML]{FE0000} \textbf{84.8}} & {\color[HTML]{FE0000} \textbf{66.8}}\\
		
		&ORTrack-D-DeiT\cite{wu2025learningorr} & CVPR 25 & {\color[HTML]{009901} \textbf{83.7}} & {\color[HTML]{FE0000} \textbf{65.1}} & {\color[HTML]{3531FF} \textbf{82.5}} & {\color[HTML]{3531FF} \textbf{59.7}} & 84.0 & {\color[HTML]{009901} \textbf{66.1}} \\ 
		
		& \cellcolor[HTML]{F2F2FF}\textbf{LGTrack-DeiT} & \multirow{-1}{*}{\cellcolor[HTML]{F2F2FF}\textbf{Ours}} & \cellcolor[HTML]{F2F2FF}{\color[HTML]{3531FF} \textbf{84.0}} & \cellcolor[HTML]{F2F2FF}{\color[HTML]{009901} \textbf{64.8}} & \cellcolor[HTML]{F2F2FF}{\color[HTML]{FE0000} \textbf{82.8}} & \cellcolor[HTML]{F2F2FF}{\color[HTML]{FE0000} \textbf{60.4}} & \cellcolor[HTML]{F2F2FF}{\color[HTML]{3531FF} \textbf{84.3}} & \cellcolor[HTML]{F2F2FF}{\color[HTML]{3531FF} \textbf{66.2}} \\ 
		\bottomrule[1pt] 
		\multirow{-7}{*}{\rotatebox{90}{ViT-based}}\\
	\end{tabular}  
	
	\setlength\tabcolsep{4.54pt}
	\begin{tabular}{ccccccccc}
		\toprule[1pt] 
		\multirow{2}{*}{} & \multirow{2}{*}{Methods} &  \multirow{2}{*}{Source} & \multicolumn{2}{c}{Avg.} & \multicolumn{2}{c}{Avg.~FPS}  & \multicolumn{1}{c}{FLOPs}  & \multicolumn{1}{c}{Param.} \\	  
		\cmidrule(lr){4-9}	
		& & & Prec. & Succ. & GPU & CPU  & GMac & M \\
		\hline
		& KCF \cite{henriques2014high} & TAPMI 15  & 52.1 & 30.0 & - & {\color[HTML]{FE0000} \textbf{624.3}} & - & - \\
		& fDSST \cite{danelljan2017discriminative} & TPAMI 17 & 59.4 & 38.2 & - & {\color[HTML]{3531FF} \textbf{193.4}} & - & - \\
		& ECO\_HC \cite{Danelljan2016ECOEC} & CVPR 17 & 67.9 & 45.3 & - & {\color[HTML]{009901} \textbf{83.5}} & - & - \\
		& AutoTrack \cite{li2020autotrack} & CVPR 20 & 70.8 & 46.6 & - & 57.8 & - & - \\
		\multirow{-5}{*}{\rotatebox{90}{DCF-based}} & RACF \cite{li2021learning} & PR 22 & 73.3 & 49.2 & - & 35.6 & - & - \\ 
		\hline
		& HiFT \cite{cao2021hift} & ICCV 21 & 74.7 & 55.3 & 160.3 & - & 7.2 & 9.9 \\
		& TCTrack \cite{Cao2022TCTrackTC} & CVPR 22 & 77.9 & 58.6 & 149.6 & - & 8.8 & 9.7 \\
		& ABDNet \cite{zuo2023adversarial} & RAL 23 & 77.2 & 58.5 & 130.2 & - & 2.8 & 32.4\\ 
		& SGDViT \cite{yao2023sgdvit} & ICRA 23 & 73.2 & 55.3 & 110.5 & - & 11.3 & 23.3 \\
		\multirow{-5}{*}{\rotatebox{90}{CNN-based}} & PRL-Track \cite{fu2024progressive} & IROS 24 & 77.2 & 57.8 & 132.3 & - & 7.4 & 12.0 \\ 
		\hline
		& LiteTrack \cite{wei2024litetrack} & ICRA 24 & 82.7 & 63.0 & 119.7 & - & - & -\\
		& SMAT \cite{gopal2024separable} & WACV 24 & 81.5 & 62.3 & 126.8 & - & 3.2 & 8.6 \\
		
		& AVTrack-DeiT \cite{wu2025learning} & ICML 24 & {\color[HTML]{FE0000} \textbf{83.7}} & {\color[HTML]{3531FF} \textbf{63.5}} & {\color[HTML]{3531FF} \textbf{260.3}} & {\color[HTML]{3531FF} \textbf{59.8}} & 0.97-1.9 & 3.5-7.9 \\
		
		&ORTrack-D-DeiT\cite{wu2025learningorr} & CVPR 25 & {\color[HTML]{009901} \textbf{83.4}} & {\color[HTML]{009901} \textbf{63.6}} & {\color[HTML]{FE0000} \textbf{292.3}} & {\color[HTML]{FE0000} \textbf{64.7}} & 1.5 & 5.3 \\ 
		
		& \cellcolor[HTML]{F2F2FF}\textbf{LGTrack-DeiT} & \multirow{-1}{*}{\cellcolor[HTML]{F2F2FF}\textbf{Ours}} & \cellcolor[HTML]{F2F2FF}{\color[HTML]{FE0000} \textbf{83.7}} & \cellcolor[HTML]{F2F2FF}{\color[HTML]{FE0000} \textbf{63.8}} & \cellcolor[HTML]{F2F2FF}{\color[HTML]{009901} \textbf{258.7}} & \cellcolor[HTML]{F2F2FF}{\color[HTML]{009901} \textbf{59.4}} & \cellcolor[HTML]{F2F2FF}2.1 & \cellcolor[HTML]{F2F2FF}7.1 \\ 
		\bottomrule[1pt] 
		\multirow{-7}{*}{\rotatebox{90}{ViT-based}}\\                              
	\end{tabular}
\end{table*}

Specifically, in terms of accuracy and success rate, LGTrack achieves an average precision (Avg.~Prec.) of 83.7\% and an average success rate (Avg.~Succ.) of 63.8\%, both of which surpass those of most comparative methods. For instance, the DCF-based tracker RACF achieves the highest precision and success rates among DCF-based methods, with average values of 73.3\% and 49.2\%, respectively, while the CNN-based tracker TCTrack performs the best among CNN-based approaches with 77.9\% and 58.6\%, yet both remain lower than LGTrack. In terms of speed, LGTrack reaches 258.7$\pm$10.0~FPS on GPU and 59.4$\pm$5.0~FPS on CPU, outperforming many real-time trackers, although KCF attains 624.3~FPS on CPU, its average precision is only 52.1\%, considerably lower than that of LGTrack. Furthermore, in terms of computational complexity, LGTrack maintains relatively low FLOPs and Params. 

Compared to other ViT-based trackers, such as AVTrack-DeiT, LGTrack-DeiT achieves the optimal Precision (Prec.) and Success (Succ.) scores when handling dense occlusions caused by parallel and intersecting vehicles in the urban traffic scenarios of the UAVDT dataset, owing to its specialized capability in occlusion representations. It also outperforms ORTrack-D-DeiT—a counterpart with dedicated optimization for occlusion challenges—in other general UAV scenarios. This verifies that the Global Grouped Coordinate Attention (GGCA) module enhances the model’s spatial feature perception and localization accuracy at an extremely low computational cost, while maintaining a high inference speed when integrated with the Similarity-Guided Layer Selection (SGLA) mechanism, which skips redundant computational layers in the Transformer.

\subsection{Ablation Experiments and Analysis} 

\noindent{(1) \textbf{Effectiveness of the ORR rule, and the SGLA and GGCA modules}}

To validate the effect of the adopted ORR rule and the SGLA and GGCA modules, we conducted a systematic ablation study on the UAVDT dataset. By doing so, the components are incrementally integrated into the baseline model to quantitatively assess their individual and combined contributions. The results are summarized in Table~\ref{tab:Backbone}.

\begin{table}[t]
	\scriptsize
	\centering
	\setlength\tabcolsep{4.0pt} 
	\renewcommand{\arraystretch}{1.2}
	\caption{Ablation Study of GGCA, ORR and SGLA in the baseline tracker.}
	\label{tab:Backbone}
	\begin{tabular}{ccccccc}
		\toprule[1pt]      
		\multirow{2}{*}{Tracker}      & \multirow{2}{*}{ORR} & \multirow{2}{*}{SGLA} & \multirow{2}{*}{GGCA} & \multicolumn{2}{c}{UAVDT}     & \multirow{2}{*}{FPS} \\
		\cmidrule(lr){5-6}
		&                      &     &                  & Prec.         & Succ.         &                      \\ \hline
		\multirow{4.5}{*}{\textbf{LGTrack-DeiT}} &            &          &                       & 78.6          & 56.7          & 226.4                \\
		& $\checkmark$                     &       &                &\textbf{}\textbf{83.4}$_{\uparrow 4.8}$ & 60.1$_{\uparrow 3.4}$ & -                    \\
		&$\checkmark$                      &$\checkmark$     &                  & 82.5$_{\uparrow 3.9}$          & 59.8$_{\uparrow 3.1}$          & \textbf{267.4}$_{\uparrow 18.1\%}$      \\  
		&$\checkmark$                      &$\checkmark$    &$\checkmark$                   & 82.8$_{\uparrow 4.2}$          & \textbf{60.4}$_{\uparrow 3.7}$          & 258.7$_{\uparrow 14.3\%}$      \\ \bottomrule[1pt] 
	\end{tabular}
\end{table}

Experimental results show that the ORR rule can significantly improve the tracking accuracy. When incorporated into the baseline tracker, the discrimination ability of the model under complex scenarios is enhanced, thus leading to consistent increases in both precision (Prec.) and success rate (Succ.). Specifically, after LGTrack-DeiT integrates ORR rules, the accuracy is improved by 4.8\% and the success rate is increased by 3.1\%. This result clearly confirms that the module effectively enhances the robustness of feature characterization of the model under occlusion interference.

In addition,  after integrating the SGLA mechanism, the tracking accuracy is basically maintained without loss while achieving significant efficiency improvement. By dynamically selecting and skipping redundant Transformer layers, the mechanism significantly optimizes the computing overhead of the model. Experiments show that after adding SGLA, the GPU speed of LGTrack-DeiT increased by 21.6\%, with only a minor decrease in performance. This proves that SGLA can trade a very small loss in accuracy for a notable gain in computational efficiency, achieving an effective balance between accuracy and speed.

\noindent{(2) \textbf{Superiority of the GGCA module}}

To validate the effectiveness of the proposed GGCA module, we compare it with the standard Coordinate Attention (CA) method\cite{hou2021coordinate}, along with other attention mechanisms such as SE\cite{hu2018squeeze} and CBAM\cite{woo2018cbam}, under the same architecture. As shown in Table~\ref{tab_effect_of_CA}, while the model with the three other attention mechanisms (SE, CBAM, and CA) already shows some improvement in precision and success rate compared to the baseline on the DTB70 test set, its performance remains significantly lower than that of the GGCA-based version. Specifically, after incorporating the GGCA module, the model achieves the best performance, with precision (Prec.) and success rate (Succ.) reaching 84.1\% and 64.8\%, respectively, corresponding to improvements of 1.8\% in precision and 1.2\% in success rate, with a loss of less than 10 FPS compared to the baseline.  

\begin{table}[!t]
	\centering
	\scriptsize
	\setlength\tabcolsep{1.5pt}
	\renewcommand{\arraystretch}{1.2}
	\caption{Ablation experiments of four different cross attention methods.}
	\label{tab_effect_of_CA}
	\begin{tabular}{cccccccc}
		\toprule[1pt]
		\multirow{2}{*}{Methods} &\multirow{2}{*}{SE\cite{hu2018squeeze}~}   &\multirow{2}{*}{CBAM\cite{woo2018cbam}~}   & \multirow{2}{*}{CA\cite{hou2021coordinate}~} & \multirow{2}{*}{GGCA} &\multicolumn{2}{c}{DTB70} &\multirow{2}{*}{FPS}\\	  
		\cmidrule(lr){6-7}	
		& & & & &  Prec. & Succ. & \\  \hline
		
		\multirow{5}{*}{\textbf{LGTrack-DeiT}} &    && &   & 82.3       & 63.6           & \textbf{267.4}\\
		&$\checkmark$ &         &                    &                            & 83.1            & 64.3       &  264.6 \\ 
		& &$\checkmark$         &                    &                                & 82.5            & 63.9       &  262.7 \\ 
		&&         & $\checkmark$                   &                                           & 83.3            & 64.1       &  262.1 \\ 
		&&         &                     & $\checkmark$                          & \textbf{84.1}$_{\uparrow 1.8}$   & \textbf{64.8}$_{\uparrow 1.2}$ & 258.7$_{\downarrow 8.7}$\\ 
		\bottomrule[1pt]     
	\end{tabular}
\end{table}

\noindent{(3) \textbf{Effectiveness of the spatial Cox-based masking operator}}

\begin{figure}[!t]
	\centering
	\includegraphics[width=1.0\textwidth]{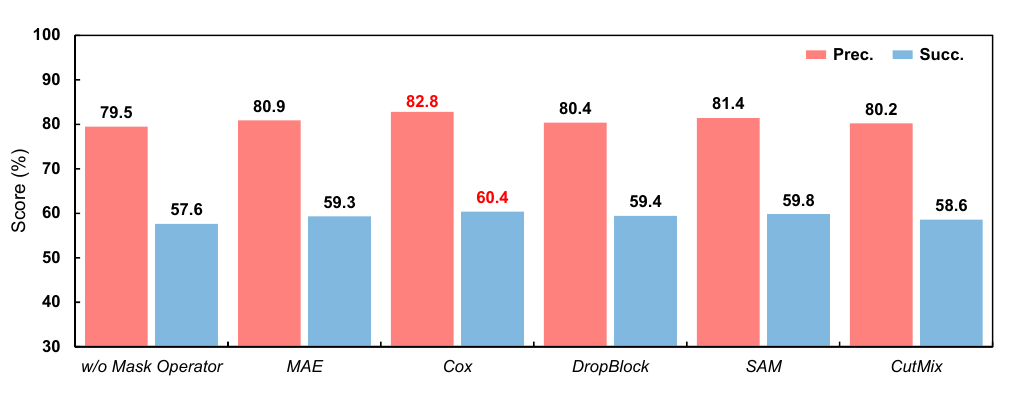}
	\caption{Ablation experiments of various masking methods on UAVDT, where the precision (Prec.) and success rate (Succ.) are used for evaluation. The configuration with Spatial Cox Processes achieves the best results.}
	\label{fig_mask}
\end{figure}

To illustrate the performance impact of various masking and representation strategies, we evaluate the model using several approaches: w/o Mask Operator, MAE\cite{He2021MaskedAA}, Cox (spatial Cox process-based masking operator), DropBlock\cite{ghiasi2018dropblock}, SAM\cite{kirillov2023segment}, and CutMix\cite{yun2019cutmix}. As shown in Figure \ref{fig_mask}, on the UAVDT\cite{du2018the}, all methods outperform the strategy without Mask Operator in terms of Precision (Prec.) and Success (Succ.). In the above strategies, the Cox method achieved the highest score, which was significantly improved compared with other methods.. These results confirm the effectiveness of the Cox-based masking operator, which yields substantially greater performance gains than existing masking and representation strategies.

\subsubsection{Sensitivity analysis of parameters}

\noindent{(1) \textbf{The configuration of saturated layers}} 

In the framework, the saturation layer is the key to balancing accuracy and speed. We use prior knowledge to directly set a fixed value for the saturation layer $l^*$, thereby simplifying and improving effectiveness. Furthermore, Table \ref{tab:Saturated} shows the impact of different configurations for saturation layers. It can be observed that the best precision-speed trade-off is achieved when $l^*$ is set to 8. We argue that when $l^*$ is set to 7, saturation is not reached in the vast majority of cases, leading to low precision and success rates due to premature stopping. Meanwhile, when $l^*$ is set to 9, there is only a 0.1\% improvement in precision at the cost of approximately 15 FPS.

\begin{table}[!t]
	\scriptsize
	\centering
	\setlength\tabcolsep{12.0pt} 
	\renewcommand{\arraystretch}{1.2}
	\caption{The impact of the configuration of saturated layers}
	\label{tab:Saturated}
	\begin{tabular}{ccccc}
		\toprule[1pt]
		\multirow{2}{*}{Methods} &  \multirow{2}{*}{Saturated layers} & \multicolumn{2}{c}{UAV123} & \multirow{2}{*}{FPS} \\	  
		\cmidrule(lr){3-4}	
		& &  Prec. & Succ. & \\
		\hline
		\multirow{3}{*}{\textbf{LGTrack-DeiT}} &      $l^*$= 7               & 81.5          & 59.2         & \textbf{265.9}                \\
		& \cellcolor[HTML]{F2F2FF}\textbf{$l^*$= 8}            &\cellcolor[HTML]{F2F2FF}\textbf{}\textbf{84.3} & \cellcolor[HTML]{F2F2FF}\textbf{66.2}                & \cellcolor[HTML]{F2F2FF}258.7                    \\
		& $l^*$= 9                         & 84.4   & 66.2          & 242.1      \\   \bottomrule[1pt] 
	\end{tabular}
\end{table}

\noindent{(2) \textbf{The configuration of the group number and pooling methods}}

\begin{figure}[!t]
	\centering
	\includegraphics[width=1.0\textwidth]{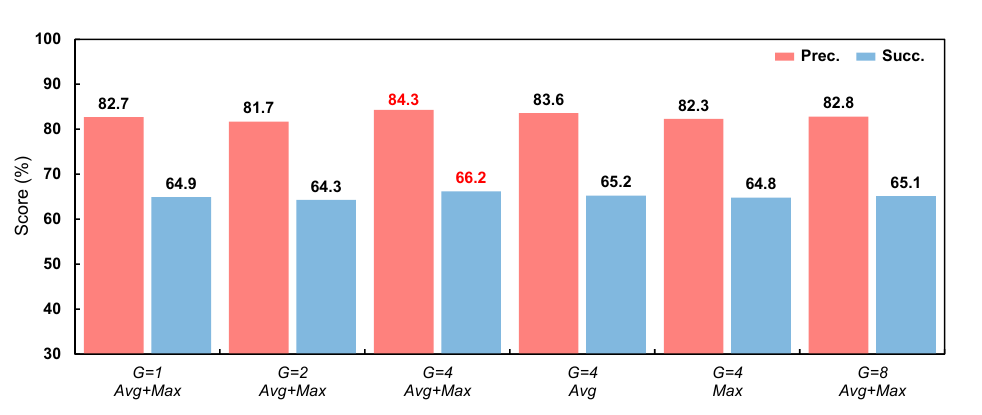}
	\caption{The impact of different numbers of groups ($G$=1, 2, 4, 8) and pooling methods (Avg, Max, Avg+Max) on UAV123. The precision (Prec.) and success rate (Succ.) are used for evaluation. The configuration with $G$=4 and Avg+Max pooling achieves the best results. }
	\label{fig_ggca}
\end{figure}

We further conduct a study on the group number and pooling methods in the GGCA. The results are shown in Figure~\ref{fig_ggca}. When the number of groups is set to $G=4$, and the combined Avg+Max pooling is adopted, the tracker achieves the best performance on the UAV123 dataset, with a precision of 84.3\% and a success rate of 66.2\%. Compared with $G=2$ and $G=8$, too few groups may weaken the modeling of spatial dependencies, while excessive groups tend to cause feature dispersion. Under the same group setting, neither average (Avg.) pooling nor max pooling alone surpasses the combination of both, which indicates that Avg+Max can capture both global distributions and salient responses simultaneously, thereby improving the robustness and accuracy of the tracker. These results indicate that GGCA, through its grouped structure and dual-path pooling mechanism, can more effectively model channel-wise and spatial dependencies, thereby achieving more robust feature enhancement in complex scenarios and validating its superiority in boosting tracking performance.

\begin{table}[t]
	\centering
	\scriptsize
	\setlength\tabcolsep{6.0pt} 
	\renewcommand{\arraystretch}{1.2}
	\caption{Application of our GGCA module to two SOTA trackers: OSTrack-256 and GRM.}
	\label{table_sota}
	\begin{tabular}{cccc}
		\toprule[1pt]
		\multirow{2}{*}{Trackers}   & \multirow{2}{*}{GGCA} & \multicolumn{2}{c}{UAVDT}\\
		\cmidrule(lr){3-4}	
		& &  Prec. & Succ.  \\
		\hline
		
		\multirow{2}{*}{OSTrack-256\cite{ye2022joint}}&                      & 85.0& 63.4\\
		& $\checkmark$                       & \textbf{86.3}$_{\uparrow 1.3}$& \textbf{64.6}$_{\uparrow 1.2}$\\ \hline
		\multirow{2}{*}{GRM\cite{gao2023generalized}}       &                      & 84.0& 62.4\\
		&  $\checkmark$                      & \textbf{85.2}$_{\uparrow 1.2}$& \textbf{63.3}$_{\uparrow 0.9}$\\ \hline
	\end{tabular}
\end{table}

\subsection{Application of GGCA on SOTA Trackers}

In order to verify the versatility and transferability of the proposed GGCA module, we integrate it into two distinct advanced tracking frameworks, OSTrack-256 \cite{ye2022joint} and GRM~\cite{gao2023generalized}, and then evaluate their performance on UAVDT \cite{du2018the}. As shown in Table~\ref{table_sota}, the performance of both trackers improved steadily across these metrics after incorporating the GGCA module. The results demonstrate that GGCA, as a versatile feature enhancement component, can be effectively embedded into different tracking architectures and consistently deliver performance gains on a general tracking dataset, further validating its broad applicability in enhancing the discriminative capability of tracking models. 

\begin{figure}[t]
	\centering
	\includegraphics[width=0.8\textwidth]{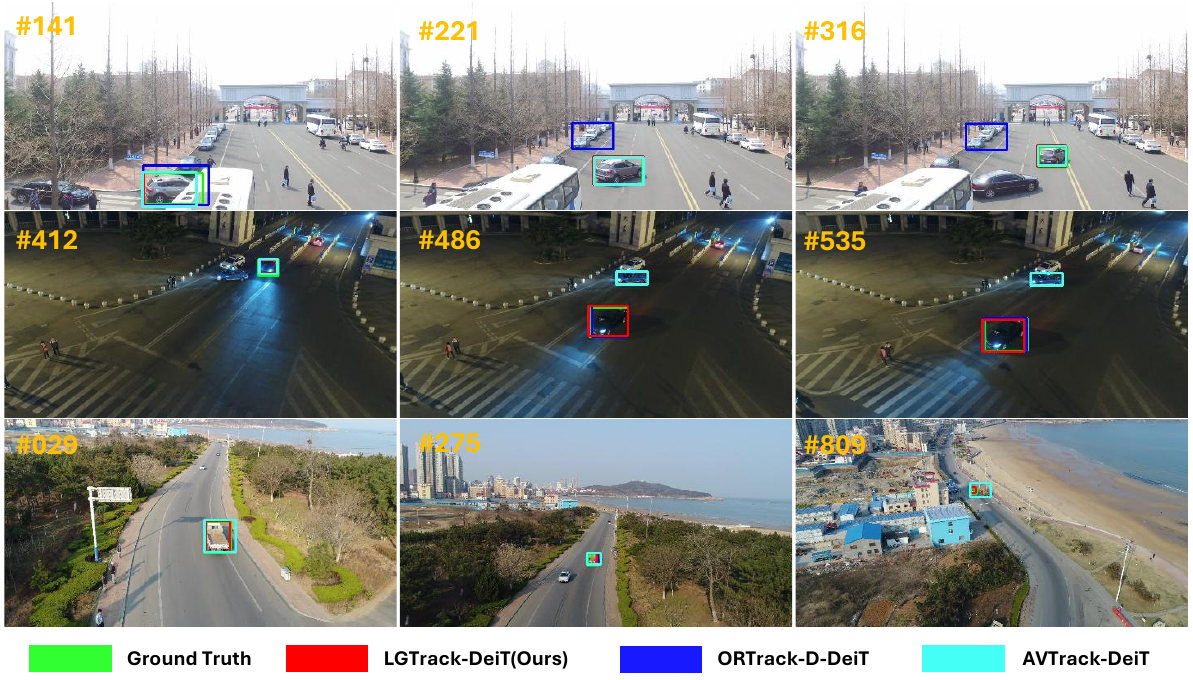}
	\caption{Qualitative evaluation on 3 video sequences from UAVDT \cite{du2018the} (i.e., S0801, S0305, and S0101).} \label{fig_bbox_visual}
\end{figure}

\subsection{Visualization}  
To intuitively demonstrate the effectiveness of the proposed method, we provide qualitative comparison results on the UAVDT \cite{du2018the} dataset. As shown in the Figure \ref{fig_bbox_visual}, under various challenging scenarios such as occlusion, scale variation, and background clutter, LGTrack-DeiT exhibits superior robustness compared to two advanced trackers, ORTrack-D-DeiT \cite{wu2025learningorr} and AVTrack-DeiT \cite{wu2025learning}. Specifically, when the target reappears after being occluded, the method in this article can quickly recapture the target and maintain stable tracking. In the case of rapid scale changes of the target, the bounding box from LGTrack-DeiT can consistently maintain a close fit to the target's contour. Even in a complex background with similar interfering objects, the model can accurately distinguish the target from interfering information and effectively avoid tracking failure. The above visual results strongly verify the effectiveness of our method.

\section{Conclusion and Further Applications }\label{sec:con} 

In this paper, an innovative tracking framework was proposed to obtain a significant improvement in efficiency by dynamically skipping redundant computational layers. At the same time, the GGCA module was introduced to enhance spatial feature perception and localization accuracy, and stochastic masking techniques were adopted to effectively improve occlusion robustness. Our method achieved state-of-the-art performance with a precision of 82.8\% and a success rate of 60.4\% on the UAVDT dataset. Experimental results also showed that our LGTrack achieved an optimal trade-off between accuracy and speed on multiple datasets. However, there is still room for improvement in handling complex scenarios like severe occlusion and rapid motion. Our future work will focus on enhancing the model's adaptability in extremely challenging situations and on developing more efficient strategies for dynamic computation allocation.

\section*{Acknowledgements}

This work was supported in part by the National Natural Science Foundation of China under Grants 62203306 and 62373251.

\section*{Declarations}

\bmhead{Data Availability Statement}  The data used in this paper are all publicly available and have no ethical violations.

\bmhead{Conflict of interest} The authors declare no potential conflict of interest.

\bmhead{Author Contributions} Yang Zhou: Conceptualization, Methodology, Writing-original draft; Derui Ding: Supervision, Writing-review \& editing, Funding;  Ran Sun: Investigation; Ying Sun: Investigation, Writing-review \& editing, Funding; and Haohua Zhang: Investigation.


\end{document}